\documentclass[10pt,twocolumn,letterpaper]{article}

\usepackage{iccv}
\usepackage{times}
\usepackage{epsfig}
\usepackage{graphicx}
\usepackage{amsmath}
\usepackage{amssymb}
\usepackage{multirow}
\usepackage{paralist}
\usepackage{makecell}
\usepackage{booktabs}
\usepackage{caption}
\usepackage{graphicx}
\usepackage{booktabs}
\usepackage{blindtext}
\usepackage{authblk}

\makeatletter

\newcommand{\Rmnum}[1]{\expandafter\@slowromancap\romannumeral #1@}
\makeatother

\usepackage[pagebackref=true,breaklinks=true,letterpaper=true,colorlinks,bookmarks=false]{hyperref}

\usepackage{color}

\iccvfinalcopy 

\def\iccvPaperID{7197} 

\ificcvfinal\fi

\begin{document}

\title{Memory-Free Generative Replay For Class-Incremental Learning}

\author{Xiaomeng Xin\textsuperscript{$\dagger$}, Yiran Zhong\textsuperscript{$\ddagger$}, Yunzhong Hou\textsuperscript{$\ddagger$}, Jinjun Wang\textsuperscript{$\dagger$}, Liang Zheng\textsuperscript{$\ddagger$}\\
 \textsuperscript{$\dagger$} Xi'an Jiaotong University\\
 \textsuperscript{$\ddagger$}Australian National University\\
   {\tt\small xmengxin@163.com, zhongyiran@gmail.com,Yunzhong.Hou@anu.edu.au, jinjun@xjtu.edu.cn, Liang.Zheng@anu.edu.au}
}

\maketitle
\ificcvfinal\thispagestyle{empty}\fi

\begin{abstract}
Regularization-based methods are beneficial to alleviate the catastrophic forgetting problem in class-incremental learning.~With the absence of old task images, they often assume that old knowledge is well preserved if the classifier produces similar output on new images. 
In this paper, we find that their effectiveness largely depends on the nature of old classes:~they work well on classes that are easily distinguishable between each other but may fail on more fine-grained ones, e.g., boy and girl.~In spirit, such methods project new data onto the feature space spanned by the weight vectors in the fully connected layer, corresponding to old classes.~The resulting projections would be similar on fine-grained old classes, and as a consequence the new classifier will gradually lose the discriminative ability on these classes.
To address this issue, we propose a memory-free generative replay strategy to preserve the fine-grained old classes characteristics by generating representative old images directly from the old classifier and combined with new data for new classifier training. To solve the homogenization problem of the generated samples, we also propose a diversity loss that maximizes Kullback–Leibler (KL) divergence between generated samples.
Our method is best complemented by prior regularization-based methods proved to be effective for easily distinguishable old classes. We validate the above design and insights on CUB-200-2011, Caltech-101, CIFAR-100 and Tiny ImageNet and show that our strategy outperforms existing memory-free methods with a clear margin. Code is available at \url{https://github.com/xmengxin/MFGR}.
\end{abstract}

\vspace{-5mm}
\section{Introduction}
Incremental learning addresses a critical problem called catastrophic forgetting:~a network often quickly forgets previously acquired knowledge when learning new knowledge~\cite{masana2020class,belouadah2020comprehensive,delange2021continual}. Between the two main categories, \ie, memory-based and memory-free methods, we choose to study the latter. Briefly, a network is asked to learn a sequence of new tasks without accessing data from previous (old) tasks. Many memory-free methods are regularization based. That is, to consolidate previous knowledge during the new knowledge learning process with extra regularization terms in the loss functions \cite{li2017learning,liumore}. 

\begin{figure}[t]
\begin{center}
 \includegraphics[width=0.45\textwidth]{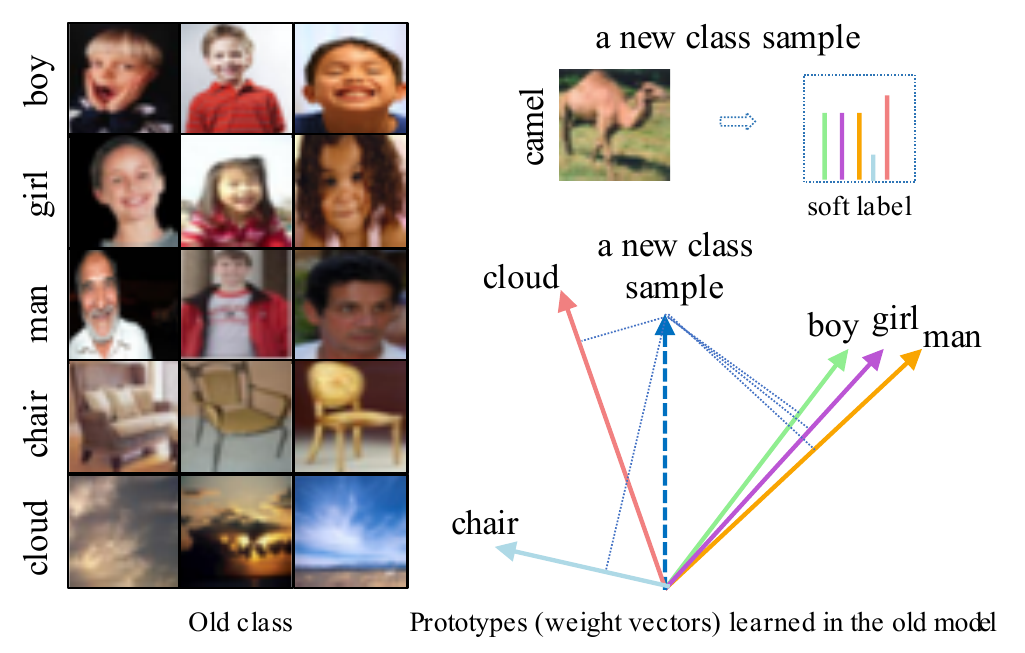}
 \end{center}
 \vspace{-6mm}
\caption{\small{A cartoon illustration of how catastrophic forgetting happens on fine-grained old classes when using regularization-based methods such as LwF \cite{li2017learning}. We draw weight vectors of five old classes and the feature vector of a new task sample. Among the five old classes, \emph{boy}, \emph{girl} and \emph{man} have similar semantics, and thus are closer in the feature space, while \emph{cloud} and \emph{chair} are further apart. When projecting the new image feature onto the weight vectors, fine-grained classes will give us nearly identical coordinates or soft labels. The new task images will require the new classifier to output similar soft labels on the three dimensions. As such, it offers little incentive for the classifier to preserve the knowledge that distinguishes between \emph{boy}, \emph{girl} and \emph{man}.}}
\vspace{-5mm}
\label{projection}
\end{figure}

In the absence of old task data, these regularization-based methods often require that the old and new classifiers give similar responses (\eg, softmax vector on old classes) to new images. This constraint benefits old knowledge preservation, as shown by the overall improvement in  old data classification. 
However, this strategy still suffers from catastrophic forgetting on \emph{old classes that are closer in the feature/semantics space}, \ie, the fine-grained classes.

Consider the cartoon example in Fig.~\ref{projection}. What the above regularization method essentially does is to project a new task image onto the weight vectors corresponding to the five classes (and ask the 5-dim projection vectors or soft labels to be similar between the new and old classifiers). In this example, given a new task image, its projections on weight vectors of the three fine-grained classes are nearly the same in value. 
In order words, in the view of new task images, weight vectors of \emph{boy}, \emph{girl} and \emph{man} carry no discriminative information among them. As a consequence, the new classifier, when using these three coordinates or soft labels as network constraints, would view them as coming from the same class; there is no longer incentive for the new classifier to distinguish them, which in effect leads to catastrophic forgetting of these fine-grained old classes.

\begin{figure}[t]
\begin{center}
 \includegraphics[width=0.45\textwidth]{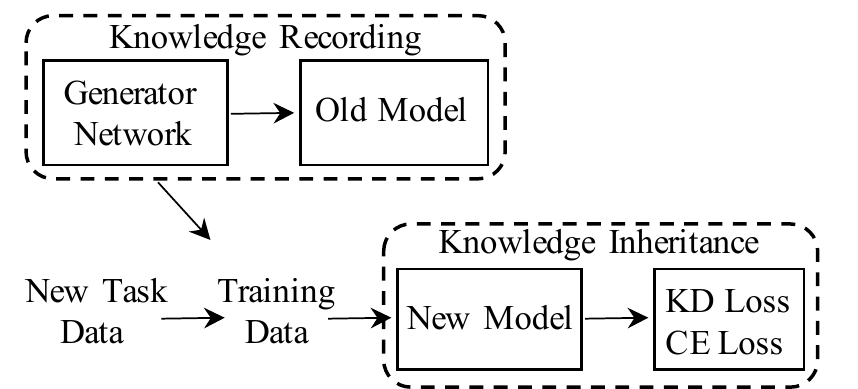}
 \end{center}
 \vspace{-5mm}
\caption{\small{System workflow. Given an old task classifier, we first generate old task images using a Generator. The generated images, together with the new task data, are used to train the new task model.}}
\label{framework}
\vspace{-5mm}
\end{figure}

In this paper, we aim to mitigate the above catastrophic forgetting problem through a generative method named memory-free generative replay. While Fig.~\ref{projection} \emph{indirectly} preserves old knowledge using new task images, we do so by \emph{directly} having the new classifier learn from the generated old task images. As shown in Fig. \ref{framework}, our framework contains two stages: knowledge recording and knowledge inheritance. Specifically, 
we first train the generator such that the generated images exhibit desired properties, \eg, a high diversity, matching the BN statistics of the old classifier, and a low domain gap with the new task data. Then, we train the new task classifier using the new task images and generated old task images that are mixed and balanced in mini-batches. Also, in practice we find that the generated samples often suffer serious homogenization problem on sample level, \ie, generating similar images for one class. To avoid this issue, we maximize Kullback–Leibler (KL) divergence between generated samples to force the samples to be diverse. 

Our method tackles the Achilles' heel of the regularization-based class-incremental learning. First, we conceptually and experimentally show that the fine-grained old classes are the most vulnerable when new tasks are learned. It is because regularization-based methods provide very little guidance for the new classifier to preserve such old knowledge. Second, although our method uses images that are much less genuine than the new target images, we still provide the characteristics of the fine-grained classes. This is proved by the significant classification accuracy improvement on these classes. 
Moreover, since the generated old images cannot perfectly cover the real data distributions, our method is best to work together with previous regularization-base methods. This validates our system: a combination of strengths of regularization-based methods and the proposed generative replay.
On standard CIFAR100 and Tiny ImageNet benchmarks, we report state-of-the-art performance among memory-free methods.

\section{Related work}
Class-incremental learning attracts more attention recently, and there are significant number of papers published on this topic~\cite{9349197}. Below we only cite a few most recent methods that closely related to our work. 

\textbf{Regularization-based Methods.}
Regularization-based methods are often memory-free, \ie, does not explicitly~\cite{rebuffi2017icarl,hou2019learning,liu2020mnemonics} or implicitly~\cite{shin2017continual,xiang2019incremental,lesort2019generative} store previous data distributions. Elastic Weight Consolidation (EWC)~\cite{kirkpatrick2017overcoming} estimates important weights through calculating diagonal approximation of the Fisher information matrix. Memory Aware Synapses (MAS)~\cite{aljundi2018memory} evaluates important parameters by looking at the sensitivity of a learned function, which can be computed in an online manner. Note that these methods require explicitly defined task id for evaluation~\cite{chaudhry2018riemannian}. Learning without forgetting (LwF) \cite{li2017learning} uses new task data to regularize the old classes outputs in new learned model. Based on it, Learning without memorizing (LwM)
\cite{dhar2019learning} introduces an attention distillation loss to regularize changes in attention maps while updating the classifier. The very recent MUC~\cite{liumore} method leverages an auxiliary classifier that is trained on an external dataset and use it to regularize new task model via multiple distillation loss. 

\textbf{Generative Replay.}
Conventional replay methods~\cite{rebuffi2017icarl} often requires explicitly storing a subset of previous samples to alleviate old knowledge forgetting. 
It limits these methods' scalability over the number of classes as it needs additional space to keep the previous samples.
Generative replay, on the other hand, does not suffer the drawbacks of conventional replay methods, and becomes an effective and general strategy for class-incremental learning~\cite{xiang2019incremental,Cong2020}. 

Recently, GAN memory~\cite{Cong2020} proposes a sequential style modulation to transfer base knowledge to new tasks simultaneously with a well-behaved GAN model to achieve no forgetting even with a long task-sequence.  
However, one remaining issue of the generative replay is that it needs to be trained on the raw samples. That means if we only provide a pre-trained model for the starting task, \ie, our problem setting, we would not be able to leverage this strategy. To tackle this issue, \cite{smith2021always} and \cite{choi2021dual} employ synthetic images that generates by data-free knowledge distillation methods~\cite{chen2019data, yin2020dreaming} to alleviate catastrophic forgetting. Comparing with them, although we also utilize such methods, we systematically show that regularization-based methods inherently suffer catastrophic forgetting on fine-grained classes. In this paper, we propose a memory-free generative replay strategy to generate representative samples for each old classes, and thus alleviating the forgetting of fine-grained classes.

\textbf{Knowledge Distillation}~\cite{hinton2015distilling} has been used in many areas such as network compression~\cite{li2020few}, semantic segmentation~\cite{liu2019structured}. However, most of these methods still require images from the original datasets. Recently, a  few work addresses data-free knowledge distillation. Chen~\etal~\cite{chen2019data} uses the classifier as a fixed discriminator and train a new generator to generate images with maximum discriminator responses. DeepInversion~\cite{yin2020dreaming} directly inverts the classifier to synthesize class-conditional input images based on the distribution of intermediate batch normalization layers. Standing on their shoulders, we adapt these data-free knowledge distillation strategies for memory-free class-incremental learning. 

\section{Preliminaries}
The aim of class-incremental learning is to prevent the catastrophic forgetting when a deep network is trying to learn a sequence of tasks. Formally, a class-incremental learning problem $\mathcal{T}$ consists of a sequence of tasks: $    \mathcal{T} = \{\mathcal{T}_1, \mathcal{T}_2, ..., \mathcal{T}_i, ..., \mathcal{T}_N\},$
where $N$ is the total number of tasks and a task $\mathcal{T}_i = (C^i, D^i), i\leq N$ consists of $k^i$ non-overlapping classes: $ C^i = \{c_1^i,c_2^i,,...,c_k^i\}, \ C^i\cap C^j=\varnothing \ \text{if} \ i\neq j,$
and the corresponding training data: $    D^i=\{(\mathbf{x}_1^i, \mathbf{y}_1^i), (\mathbf{x}_2^i, \mathbf{y}_2^i), ..., (\mathbf{x}_k^i, \mathbf{y}_k^i)\},$
where $\mathbf{x}_k^i$ are the input images of class $c_k^i$ and $\mathbf{y}_k^i$ is one-hot ground truth label. The goal of class-incremental learning is to minimize the classification error $\ell_\text{class}$ among all observed tasks $\mathcal{T}$:
\begin{equation}
    \min \frac{1}{N}\sum_{i=1}^N \ell_\text{class}[\mathcal{F}^N(D^i;\theta^N), \mathbf{y}^i],
\end{equation}
where $\mathcal{F}^N$ is the classification network for $\mathcal{T}$ and $\theta^N$ is its weights.

In memory-free class-incremental learning setup~\cite{dhar2019learning}, the only available information for learning task $\mathcal{T}_i$ is the classification network of the previous task $\mathcal{F}^{i-1}(\cdot)$ and the current training data $D^i$.
For the simplicity of notations, we omit the task number and only consider two consecutive tasks (an old task and a new task) for the rest of this paper.

\section{Revisiting Regularization-based methods}

Let us further divide the weight $\theta$ of a classification network into three groups: 1) a set of shared parameters $\theta_s$, \ie, the weight of feature extraction layers; 2) the task-specific parameters for the old task $\theta_{o}$,  \ie, the weights of the fully-connected layer that connected to the old task output softmax layer, and 3) the task-specific parameters for the new task $\theta_{n}$. As the learning is continuous, the $\theta_{n}$ will be added to $\theta_{o}$ to classify more classes.

We use LwF~\cite{li2017learning} as a representative of regularization-based methods. There are two losses used in LwF~\cite{li2017learning}, one for training the classification network on new task and the other for maintaining the accuracy on old tasks. The former is a standard multinomial logistic loss~\cite{krizhevsky2012imagenet} between the network output of the new task $\hat{\mathbf{y}}^{n}$ and the one-hot ground truth label $\mathbf{y}^{n}$:
\begin{equation}
    \ell_{\text{ce}}(\mathbf{y}^{n},\hat{\mathbf{y}}^{n} ) = -\mathcal{H}(\mathbf{y}^{n}, \hat{\mathbf{y}}^{n}),
\end{equation}
where $\mathcal{H}(\cdot)$ represents the cross-entropy loss.

The latter is a Knowledge Distillation loss~\cite{hinton2015distilling} that maintains old knowledge using new data:
\begin{equation}
    \ell_{\text{nkd}}(\mathbf{y}^{o},\hat{\mathbf{y}}^{o} ) = -\mathcal{H}(\mathbf{y}^{ o},\hat{\mathbf{y}}^{o}),
\end{equation}
where $\mathbf{y}^{o}$ is the output of the old classification network on the new data and $\hat{\mathbf{y}}^{o}$ is the $\theta_{o}$ output of the new network on the new data. The $\mathbf{y}^{o}, \hat{\mathbf{y}}^{o}$ are often scaled with a temperature $T$ for better emphasis on smaller probabilities~\cite{li2017learning}.

Note that the $\ell_{\text{nkd}}$ implies an assumption that if two networks have same output for new task data, the old knowledge will be effectively consolidated. Here, we show that this assumption may not hold in classification task which has many hardly distinguishable classes, such as fine-grained classes, as the $\ell_{\text{nkd}}$ ability largely depends on the inter-class distances between the old classes. 

To be specific, the old knowledge is encoded as the probabilities on each old class by projecting new data onto the feature space spanned by the old model. If the inter-class feature distance between two old classes is similar e.g., boy and girl, the probabilities of new data on these classes would be similar. Such supervised label will make the new model gradually lose its ability to discriminate these classes. 
\begin{table*}[t]
 \begin{minipage}{0.65\linewidth}
  \centering
  \includegraphics[width=\linewidth, trim=0 10mm 0 10mm]{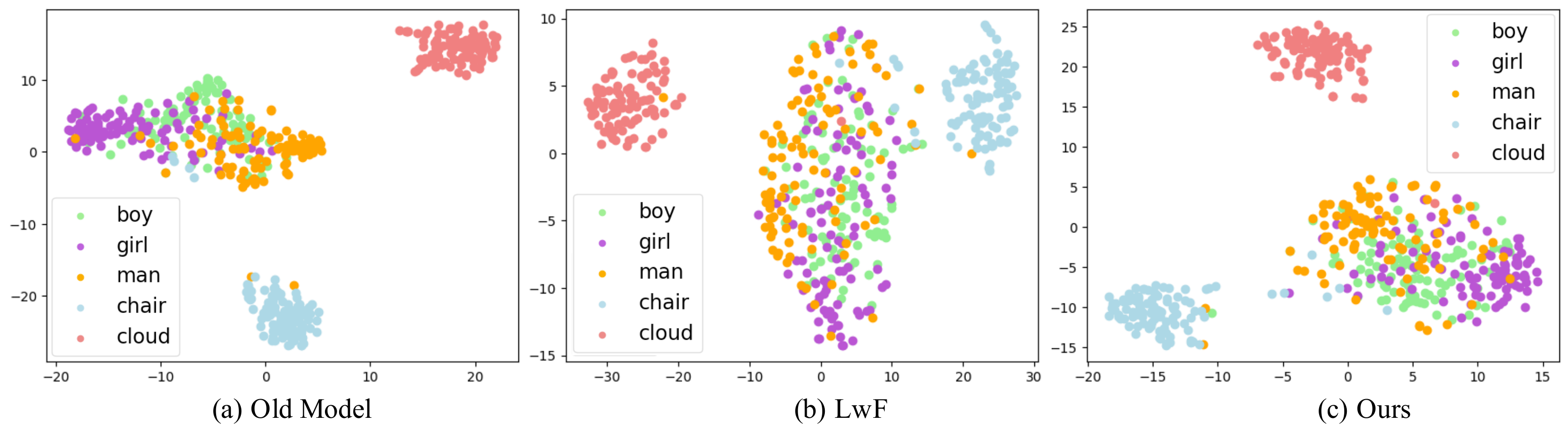}
  \captionof{figure}{\small{T-SNE visualization of image from five old classes on (a) the old task model, (b) LwF, and (c) our method. As expected, fine-grained classes \emph{boy}, \emph{girl} and \emph{man} are very close in the old task model and become highly mixed with LwF. In our system, these classes are better separated than LwF, while ``easy'' classes are still well clustered.}}
  \label{tsne}
 \end{minipage}
 \hfill
  \begin{minipage}{0.32\linewidth}
  \centering
  \includegraphics[width=\linewidth,  trim=0 3mm 0 3mm]{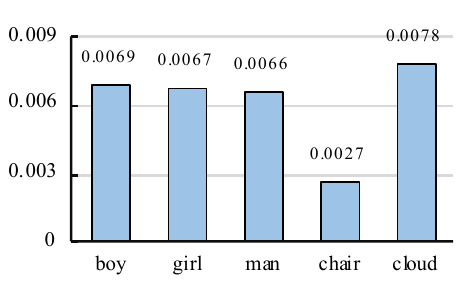}
  \captionof{figure}{\small{Avg.~projected probabilities on five old classes. For each class, we average the softmax values from new task images. We can observe that fine-grained classes have similar probabilities.}}
  \label{softmax}
 \end{minipage}
 \hfill
 \vspace{-3mm}
\end{table*}

To verify our hypothesis, we visualize T-SNE (Fig.~\ref{tsne}) and show the projected probabilities (Fig.~\ref{softmax}) of five old classes. Among them, three classes (\emph{boy}, \emph{girl}, and \emph{man}) are similar to fine-grained classes which have close projected probabilities, the other two are normal classes (\emph{chair}, \emph{cloud}) that are easy to distinguish.
Compared with the two normal classes, we can see that the three fine-grained classes have lower inter-class distances in the old model feature space and the distance gets lower in LwF (Fig.~\ref{tsne}). We also can observe that LwF indeed suffers severe accuracy drop on these fine-grained classes (Table~\ref{top_1}).

\begin{table}[t]
\scriptsize
\renewcommand\arraystretch{1.15}
  \begin{center}
    \caption{\small{Top-1 Accuracy of different models on selected classes.}}
    \label{top_1}
    \vspace{-3mm}
    \setlength{\tabcolsep}{3.8mm}
      \begin{tabular}{c |c |c| c| c| c}
      \hline
      Top-1 & boy & girl & man & chair & cloud  \\
      \hline
      \hline
      Old Model &  55\% & 69\% & 60\% & 87\% & 96\%\\
      \hline
      LwF & 18\% & 42\% & 41\% & 80\% & 89\%\\
      \hline
      Ours & 38\% & 60\% & 56\% & 81\% & 92\%\\
      \hline
      \end{tabular}
  \end{center}
  \vspace{-7mm}
\end{table}

This is an inherent limitation of regularization-based methods, can we avoid it in the memory-free incremental learning setup, \ie, maintaining the characteristics of old classes without using the old task data? In the following section, we show that it can be achieved by the proposed memory-free generative replay strategy.

\section{Proposed Framework}
In this section, we start with the overall workflow of our framework and then provide details for each component in the following subsections.

\subsection{Overall Workflow}
Figure~\ref{framework} illustrates the overall workflow of our framework. To learn a new task $\mathcal{T}_n$, we have two sequential learning stages: 1) knowledge recording; 2) knowledge inheritance. At the first stage, a generative model is trained to recall old image knowledge only from the old classifier; Then at the second stage, the generative model generates old task samples on the fly to train a new task classifier together with the new task samples. 

Recent literature has shown that a deep network model can be compressed without original real training data~\cite{chen2019data,yin2020dreaming}. However, it is non-trivial to adapt this data-free knowledge distillation strategy~\cite{chen2019data} to class-incremental learning. There are several requirements for the generated samples: 
1) can produce similar response as the real old samples on the old classifier; 
2) have small domain gap between the generated samples and the old real samples; 
3) have enough inter-class and intra-class divergence between samples to avoid homogenization. For these requirements, only the first part can be solved by DAFL~\cite{chen2019data}, and we use it as our baseline method. In this paper, we adopt the batch normalization loss~\cite{cai2020zeroq} to reduce the domain gap and propose a diversity loss to address the homogenization problem. More technical details will be explained in the sequel of this paper.

\subsection{The Generative Model}

\textbf{Baseline.}
Given an old task classification network $\mathcal{F}^o$, we train a generator $\mathbb{G}$ to generate samples that follow the old task data distribution.

To achieve this goal, we assume that if a sample has similar output as the old task data on the old model, it should belong to the old data distribution. In this case, instead of mimic the old data, we require $\mathbb{G}$ to mimic the original old data output on $\mathcal{F}^o$, \ie one-hot like vectors. 
Specifically, let us denote $n$ generated samples for the old task as $\widetilde{\mathbf{X}}^o=\{\tilde{\mathbf{x}}_1^o,\tilde{\mathbf{x}}_2^o,...,\tilde{\mathbf{x}}_n^o\}$, and the corresponding outputs of $\mathcal{F}^o$ are $\widetilde{\mathbf{Y}}^o=\{\tilde{\mathbf{y}}_1^o,\tilde{\mathbf{y}}_2^o,...,\tilde{\mathbf{y}}_n^o\}$, where $\tilde{\mathbf{y}}^o = \mathcal{F}^o(\tilde{\mathbf{x}}^o)$. To force $\tilde{\mathbf{y}}^o$ to be a one-hot like vectors, we generate a pseudo one-hot vector $\bar{\mathbf{y}}^o = \arg \text{max}(\tilde{\mathbf{y}}^o)$ and minimize the cross-entropy between $\tilde{\mathbf{y}}^o$ and $\bar{\mathbf{y}}^o$ over all outputs: $\ell_{oh}= -\frac{1}{n}\sum_{l=1}^n\mathcal{H}(\bar{\mathbf{y}}_l^o,\tilde{\mathbf{y}}_l^o).$
It is possible that the generator would only generate one or few class samples. Therefore, to avoid this homogenization situation, we introduce a class-diversity loss~\cite{chen2019data} $\ell_\text{cd} = -H(\frac{1}{n}\sum_{l=1}^n\widetilde{\mathbf{y}}_l^o)$ to balance the classes in generated images, where $H(\cdot)$ denotes the information entropy.
This loss will reach its minimum if the classes of generated samples are evenly distributed (see \cite{chen2019data} for more details).
Our final baseline loss is 
\begin{equation}
    \ell_{\text{base}} = \ell_{oh} +\lambda_1\ell_{\text{cd}}
\end{equation}

\begin{figure}[t]
\begin{center}
 \includegraphics[width=\columnwidth,trim=0 12mm 0 6mm]{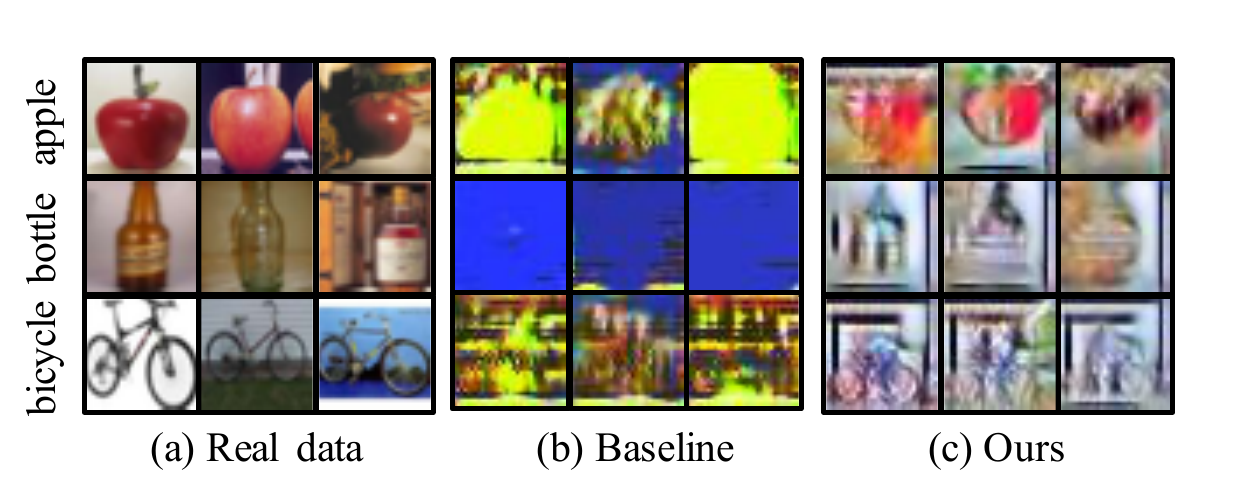}
 \end{center}
\caption{\small{Visualization of generated samples. Our generated images (c) look more similar to real data (a) than the baseline (b). They also exhibit good diversity.}}
 \vspace{-6mm}
\label{G_visualization}
\end{figure}

\textbf{Alignment with batch normalization statistics.}
In order to reduce the domain gap between the generated and real old samples, we introduce a loss that aligns batch normalization (BN) statistics.
Here we first simply assume that the domain gap between two datasets can be reflected by their statistic distribution, \ie, mean $\mu$ and variance $\sigma^2$ if we assume the data follow the Gaussian distribution. Based on this assumption, we require the generated samples to have similar mean $\mu$ and variance $\sigma^2$ to the old real samples. However, since the old real samples is unavailable, we cannot get old real data mean and variance. Instead,
we employ batch norm statistics which are stored in batch normalization layers of the old model to approximate the old real data statistics~\cite{yin2020dreaming}. Our batch normalization loss is
\begin{equation}
    \ell_\text{bn} = \sum\left|\left|\mu(\tilde{\mathbf{x}}^o) - \mu_{bn}\right|\right|_2 + \sum\left|\left|\sigma^2(\tilde{\mathbf{x}}^o) - \sigma^2_{bn} \right|\right|_2.
\end{equation}
An example of generated samples with the batch normalization loss is shown in Fig.~\ref{G_visualization}.

\textbf{Sample diversification.}
With $\ell_\text{cd}, \ell_\text{bn}$, the generator can generate samples that are evenly distributed in old task classes and match the old data distributions, but it may still face homogenization at sample level, \ie, generate similar samples for the same class, and thus limit the performance. 

To avoid this situation, we propose a diversity loss that maximizes Kullback–Leibler (KL) divergence between two generated samples and thus increase the diversity. Formally, given two generated samples $\tilde{\mathbf{x}}_1^o, \tilde{\mathbf{x}}^o_2$, our diversity loss for a pair of samples ($\tilde{\mathbf{x}}_1^o, \tilde{\mathbf{x}}^o_2$) is:
\begin{align}
    \ell_\text{div} &= - \frac{1}{2}(\mathcal{D}_{\text{KL}}(\tilde{\mathbf{x}}_1^o||\tilde{\mathbf{x}}_2^o) + \mathcal{D}_{\text{KL}}(\tilde{\mathbf{x}}_2^o||\tilde{\mathbf{x}}_1^o)),
\end{align}
where $\mathcal{D}_{\text{KL}}$ is the Kullback–Leibler (KL) divergence between two generated samples. For computational efficiency, we randomly select $200$ pairs of the generated samples and compute the diversity loss among them.

\subsection{Training procedure}
\textbf{Knowledge recording}
We use a generator to recall the old knowledge from the old task classifier. The overall loss functions for the generator are:
\begin{equation}
\label{stage_one_loss}
    \ell_{\text{KR}} = \ell_{base}+\lambda_2\ell_{\text{bn}} +\lambda_3\ell_{\text{div}},
\end{equation}

\textbf{Knowledge inheritance}
Once the generator for old tasks is trained, we use it to generate pseudo samples of the old tasks and utilize them to train the new task classifier. Our pseudo samples are generated in an online fashion and for each batch, we match the number of samples for each class to avoid class-imbalance problem. For the new task samples, we use the cross entropy loss $\ell_{\text{ce}}$ and for the pseudo old task samples, we use the knowledge distillation loss $\ell_{\text{gkd}}$ as we do not know their corresponding classes. Since the generated samples provide complementary information for LwF~\cite{li2017learning}, We add the $\ell_{\text{nkd}}$ loss in LwF~\cite{li2017learning} to our overall loss functions:
\begin{equation}
\label{stage_two_loss}
    \ell_{\text{KI}} = (1 - \lambda_4)\ell_{\text{ce}} +  \lambda_4(\ell_{\text{gkd}} + \ell_{\text{nkd}}),
\end{equation}

\subsection{Discussion}
\textbf{Why is LwF complementary to our design?} In our system, the inclusion of the LwF losses is necessary for two reasons. First, LwF is effective in preserving the discriminative abilities of the old classifier for easily distinguished classes while weak in fine-grained classes as shown in Table~\ref{top_1}, Fig. \ref{projection} as well as our ablation studies (Table \ref{Ablation}). Second, while our method generates higher-quality images than the baseline (Fig. \ref{G_visualization}), we still notice a large gap between these images and the real ones. In comparison, LwF uses new target images which do not have such a domain gap.

\section{Experiment}
\subsection{Datasets}
We conduct our experiments on four datasets: CUB-200-2011~\cite{wah2011caltech}, Caltech-101~\cite{fei2004learning}, CIFAR-100~\cite{krizhevsky2009learning}, and Tiny ImageNet~\cite{le2015tiny}.
\textbf{CUB-200-2011} contains 200 fine-grained bird classes and a total number of 11788 images. 
\textbf{Caltech-101} has 101 categories. Each category in it has about 40 to 800 images and most categories have about 50 images. The size of each image is roughly $300\times200$ pixels.
\textbf{CIFAR-100} consists of 60000 $32\times32$ colour images in 100 classes. There are 500 training images and 100 testing images per class.
\textbf{Tiny ImageNet} contains 200 classes of images with a size of $64\times64$ for training and each class has 500 training images, 50 validation images and 50 testing images. Since the category label of the testing images is not available, we estimate the performance on the validation set~\cite{liumore}. 
\begin{table*}[htb!]
\renewcommand\arraystretch{1.05}
  \begin{center}
    \caption{Ablation study of various loss functions (defined in Eq.~\ref{stage_one_loss}, Eq.~\ref{stage_two_loss}) on CIFAR-100 and Tiny ImageNet. Results indicate that these loss objectives are indispensable in our system.}
    \vspace{-3mm}
    \label{Ablation}
    \resizebox{\textwidth}{!}{
      \begin{tabular}{l|ccc|ccc|ccccc|ccccc}
      \hline
      \multirow{2}{*}{Method} & \multicolumn{3}{c|}{Stage~\Rmnum{1}} & \multicolumn{3}{c|}{Stage~\Rmnum{2}} & \multicolumn{5}{c|}{CIFAR-100} & \multicolumn{5}{c}{Tiny ImageNet} \\
      \cline{2-17}
      \multirow{2}{*} & $\ell_{\text{base}}$ & $\ell_{\text{bn}}$ & $\ell_{\text{div}}$ & $\ell_{\text{gkd}}$ & $\ell_{\text{nkd}}$ & $\ell_{\text{ce}}$ & Task 1 & Task 2 & Task 3 & Task 4 & Task 5 &  Task 1 & Task 2 & Task 3 & Task 4 & Task 5 \\
      \hline
      Fintune & & & & &  & \checkmark & 84.20 & 44.42 & 29.27 & 21.27 & 17.98 & 65.40 & 35.73 & 23.68 & 16.14 & 13.96 \\
      \hline
      LwF & & & & & \checkmark & \checkmark & 84.20 & 70.83 & 61.20 & 51.83 & 44.96 & 65.40 & 53.40 & 42.17 & 32.94 & 25.70 \\
      \hline
      LwF  w. $G_{base}$ & \checkmark & & & \checkmark & \checkmark & \checkmark & 84.20 & 70.40 & 59.42 & 52.21 & 46.39 & 65.40 & 51.35 & 39.63 & 32.30 & 28.61 \\
      \hline
      $G_{bn}$ & \checkmark & \checkmark & & \checkmark & & \checkmark & 84.20 & 54.38 & 40.63 & 32.26 & 27.10 & 65.40 & 46.15 & 31.07 & 20.05 & 15.80 \\
      \hline
      LwF w. $G_{bn}$ & \checkmark & \checkmark & & \checkmark & \checkmark & \checkmark & 84.20 & 71.40 & 62.17 & 54.30 & 49.06 & 65.40 & 54.95 & 45.17 & 37.05 & 32.71 \\
      \hline
      LwF w. $G_{bn+div}$ & \checkmark & \checkmark & \checkmark & \checkmark & \checkmark & \checkmark & 84.20 & \textbf{71.12} & \textbf{62.25} & \textbf{54.75} & \textbf{49.73} & 65.40 & \textbf{56.83} & \textbf{48.17} & \textbf{39.35} & \textbf{34.46} \\
      \hline
      \end{tabular}
      }
  \end{center}
  \vspace{-3mm}
\end{table*}

\subsection{Implementation Details}
We implement our model with Pytorch. 
For CUB-200-2011 and Caltech-101, we design a new generator structure based on DCGAN~\cite{dcgan} generator that can generate high resolution images, e.g., $256\times256$. ResNet18~\cite{he2016deep} is employed as the classifier, and all the training images in knowledge inheritance stage are resized to $224\times224$. 
For CIFAR-100 and Tiny ImageNet, the generator from DCGAN and ResNet34~\cite{he2016deep} are adopted as our generator and the classification backbone respectively. 

For the initial non-incremental state, we train the model with a mini batch size of 128, an initial learning rate of 0.01 and stop training at epoch 150.
For the subsequent incremental states, we first train the generator for 500 epochs with a learning rate of 0.01 and a mini batch size of 512 in each incremental step.
$\lambda_1$ is set to 5 according to~\cite{chen2019data}, $\lambda_2$ and $\lambda_3$ are set to 20 and 0.1 respectively, and we put hyperparameter analysis into our supplementary material.
New task model is finetuned on old task model with generated old data and new training data. The ratio between the images number of each class (old and new) in a minibatch is 1:1. We set $\lambda_4$ to iteration~/~(iteration + 1) which is adaptive to the iteration of the incremental step. The initial learning rate is set to 0.01 and we train each classification model for 150 epochs.
\vspace{-2mm}
\subsection{Method Analysis}

\textbf{The effectiveness on fine-grained classes.}
In order to better verify our hypothesis that LwF~\cite{li2017learning} fails to consolidate fine-grained classes knowledge while the proposed framework can effectively improve accuracy on old fine-grained classes, we design a setting and conduct it on CUB-200-2011 and Caltech-101. 
In our setting, there are two tasks: old and new. We randomly select 40 fine-grained classes from CUB-200-2011 and 10 normal classes from Caltech-101 as old task classes, and choose another 10 classes from Caltech-101 as new task classes. 
In training stage, for reducing the interference of other possible factors, such as unbalanced number of images in each class, we arrange this number to about 30 for each class for both training and testing data.

Table~\ref{fine_grained} shows the average Top-1 accuracy of 40 fine-grained and 10 normal classes respectively.
Comparing with Oracle which is the upper bound, we can see that the average class Top-1 of LwF on 10 normal classes slightly decreases while it sharply decrease by $9.61\%$ on fine-grained classes. This further verify our hypothesis that the ability of LwF largely depends on the inter-class distance between the old classes, and illustrate that forgetting in LwF mainly on fine-grained classes which has similar inter-class feature distance.
Compared with LwF, our method not only maintain the Top-1 on normal classes but also increase the old fine-grained class accuracy by $5.99\%$.  

\begin{table}[t]
\scriptsize
\renewcommand\arraystretch{1.15}
  \begin{center}
    \caption{\small{Average class Top-1 accuracy on CUB-200-2011 and Caltech-101 under our 2-phase setting. This verifies our hypothesis that the ability of LwF largely depends on the inter-class distance between the old classes and that the generated images better consolidate old fine-grained classes knowledge.}}
    \vspace{-3mm}
    \label{fine_grained}
    \setlength{\tabcolsep}{1.9mm}{
      \begin{tabular}{l|ccc|c|c}
      \hline
      \multirow{3}{*}{Method} & \multicolumn{5}{c}{Task 1 (new task)} \\
      \cline{2-6}
      & \multicolumn{3}{c|}{Old classes} & New classes & Total \\
      \cline{2-6}
      & 40 fine-grained & 10 normal & 50 old & 10 new & 60 total \\
      \hline
      Oracle & 79.90 & 96.61 & 83.20 & 98.09 & 85.68 \\
      \hline
      LwF~\cite{li2017learning} & 70.29 & 95.07 & 75.25 & 92.93 & 78.20\\
      \hline
      Ours & \textbf{76.28} & 95.01 & 80.02 & 91.39 & 81.92\\
      \hline
      \end{tabular}
      }
  \end{center}
  \vspace{-5mm}
\end{table}

\textbf{Ablation study of different losses.}
We employ 5-phase setting on CIFAR-100 and Tiny ImagNet to analyse the contribution of our different losses. In this setting, each dataset are split into five tasks that has equal classes number. Table~\ref{Ablation} shows the comparison results.
From Table~\ref{Ablation}, we can see that LwF outperforms Finetune by a large margin, and this shows that using new data to preserve old knowledge is somehow effective.
$G_{base}$ is the data generated by the generator trained with $\ell_{\text{base}} = \ell_{oh} +\lambda_1\ell_{\text{cd}} $. Using both new data and $G_{base}$ to train new task,
the performance on CIFAR-100 and Tiny ImageNet is a slight increase in some tasks, and a minor decrease in another tasks. We argue that this is because there exists domain gap between $G_{base}$ and old task data. When we introduce $G_{bn}$ to regularize new model with LwF, we can see all the tasks performance are improved. In addition, combined with $\ell_{\text{div}}$, our method achieves the best performance.

\begin{table}[t]
\scriptsize
\renewcommand\arraystretch{1.15}
  \begin{center}
    \caption{\small{Average class Top-1 accuracy on CUB-200-2011 under 2-phase setting: 10 classes for old task and 10 classes for new task.}}
    \vspace{-3mm}
    \label{deepinversion}
    \setlength{\tabcolsep}{5.8mm}{
      \begin{tabular}{l|cc|c}
      \hline
      \multirow{2}{*}{Method} & \multicolumn{3}{c}{Task 1 (new task)} \\
      \cline{2-4}
      & Old & New & Total \\
      \hline
      Oracle & 81.45 & 92.46 & 86.96 \\
      \hline
      LwF~\cite{li2017learning} & 54.31 & 92.11 & 73.21 \\
      \hline
      DeepInversion~\cite{yin2020dreaming} & 66.83 & 81.86 & 74.34 \\
      \hline
      Ours & 70.94 & 85.59 & 78.27 \\
      \hline
      \end{tabular}
      }
  \end{center}
  \vspace{-5mm}
\end{table}

\textbf{Computational cost.} Our method involves a generation stage and a discriminative learning stage. Specifically, on the CIFAR-100 and a GeForce RTX 3090 GPU, we spend 5 hours in training the generator once. Since we generate old task images on the fly with discriminative learning, our classification process takes 1 hours to train once.

\textbf{Using other data-free knowledge distillation method.} We first design a 2-phase setting: 20 classes (10 classes for old task and the rest 10 classes for new task ) are randomly selected form CUB-200-2011 and conduct experiments on it.
DeepInversion~\cite{yin2020dreaming}, which is a popular data-free knowledge distillation method, is introduced to generate synthetic images. From Table~\ref{deepinversion}, we can see that both Deepinversion and Ours can alleviate old knowledge forgetting and this illustrates using other method to generate pseudo samples also work. Here the advantage of Ours is that our method is in fact more time efficient than DeepInversion. Specifically, the training time for our generator is fixed, and generating new images is very fast. 
In comparison, while Deepinversion does not need to train a generator, it has to do gradient computation whenever generating a new image. This process is time consuming especially when considering new tasks keep coming in. 

\begin{figure}[!t]
\begin{center}
 \includegraphics[width=0.4\textwidth]{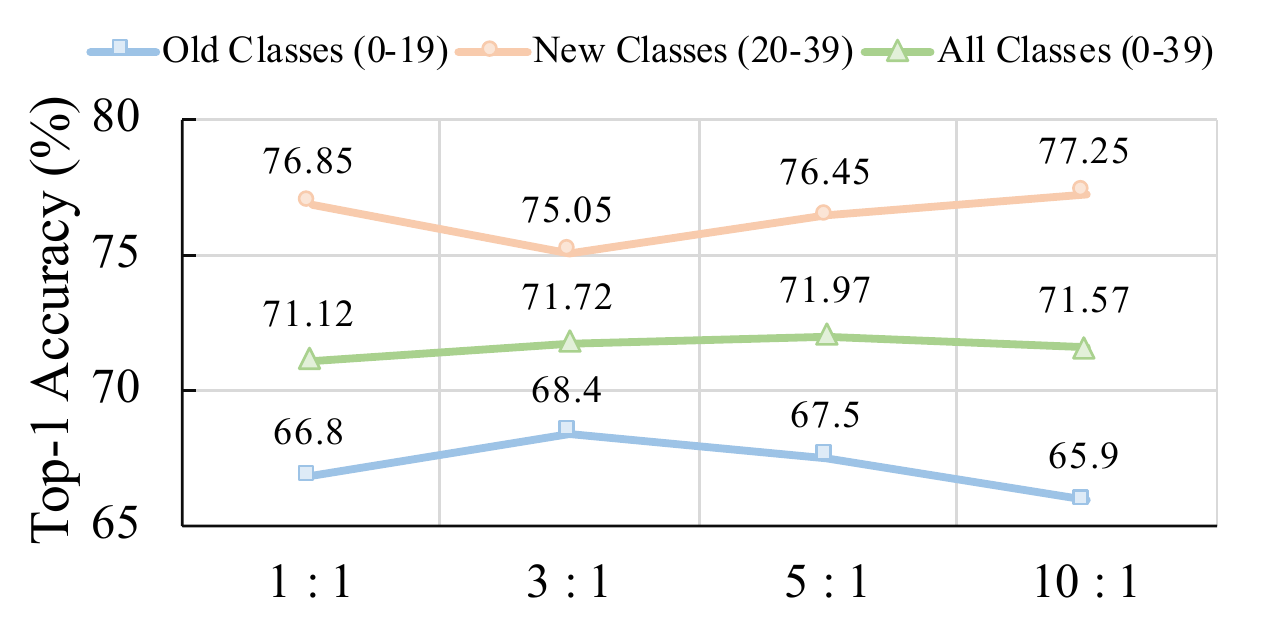}
 \end{center}
 \vspace{-5mm}
\caption{{Impact of the ratio between numbers of generated old data and new data from CIFAR-100. The average accuracy of the 20 old classes, new classes (20-39) and all classes are shown.}}
\label{dif_ratio_G}
\end{figure}

\textbf{The impact of the number of generated images.}
We use different ratios of generated data and new data (G:N) in a minibatch when training the new model. The four different ratios are set to 1:1, 3:1, 5:1, and 10:1. 
We conduct experiments on CIFAR-100 in 5 phases setting.
Fig.~\ref{dif_ratio_G} reports the top-1 accuracy of task 2 on old classes (0-19), new classes (20-39), and all seen classes (0-39) with four different ratios.
From Fig.~\ref{dif_ratio_G}, we observe that the top-1 accuracy of all seen classes in different ratios are close, it seems that the number of generated images in a minibatch is not sensitive when learning new incremental task.

\begin{figure}[!t]
\begin{center}
 \includegraphics[width=0.4\textwidth]{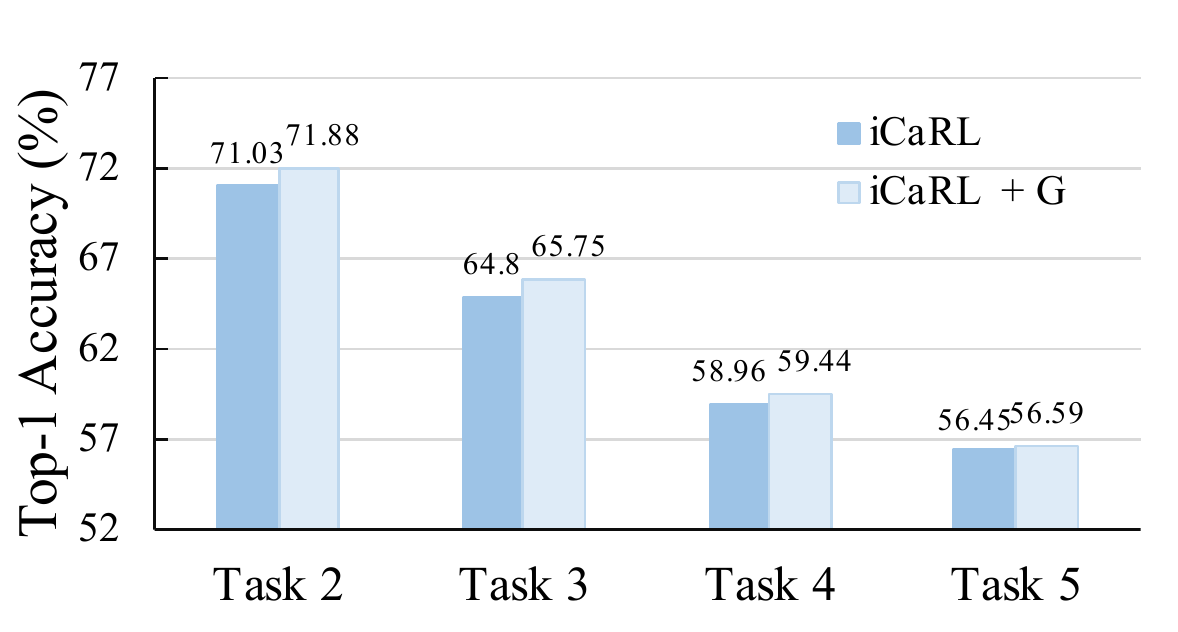}
 \end{center}
 \vspace{-5mm}
\caption{Classification accuracy when applying our method to the memory-based method \cite{rebuffi2017icarl} on CIFAR-100. We show that the generated samples bring slight benefit.}
\label{iCaRL_G}
\end{figure}

\textbf{Can our method be combined with memory-based systems?}
We apply our generated data to one representative memory-based method iCaRL~\cite{rebuffi2017icarl} to see whether the generated data is help for memory-based method. We conduct the experiment in 5 phases setting on CIFAR-100. 
For iCaRL,  we run the official released code with a fixed capacity (2000 exemplars). 
Fig.~\ref{iCaRL_G} shows that our generated data do not decrease the top-1 accuracy of iCaRL but with a slightly improvement. 
This might be caused by the domain gap between old raw data and generated data.  We would like to further analyse it in the future.

\textbf{Relationship between our method and memory-based ones.} Our system and memory-based methods share an advantage, \ie, preventing forgetting by \emph{directly} replaying the (generated) old data when training the new task classifier. The difference is also apparent: our generated images are not real, and thus need the complementarity from regularization-based methods. Besides, as to be shown in the experiment, it is interesting to see that our method slightly improves a memory-based method \cite{rebuffi2017icarl}. It suggests that the generated images may add some useful cues to the system such as high diversity. A complete study of this phenomenon is out of the scope of this paper, and we will leave it to future work.

\begin{table}[t]
\scriptsize
\renewcommand\arraystretch{1.15}
  \begin{center}
    \caption{\small{Method comparison on CIFAR-100 under the 5-phase setting. Top-1 classification accuracy is shown. The rightmost column shows the improvement of our method over the corresponding one, computed on all the tasks.}}
    \vspace{-3mm}
    \label{5p_cifar}
    \setlength{\tabcolsep}{2.3mm}{
      \begin{tabular}{l|llll|c}
      \hline
      \multirow{2}{*}{Method} & \multicolumn{4}{c|}{CIFAR-100} & \multirow{2}{*}{\shortstack{All task avg.\\improvement}} \\
      \cline{2-5}
      \multirow{2}{*}{} & Task 2 & Task 3 & Task 4 & Task 5 & \multirow{2}{*}{} \\
      \hline
      Oracle & 82.38 & 81.45 & 78.46 & 79.67 & -\\
      \hline
      MAS~\cite{aljundi2018memory} & 47.17 & 30.78 & 23.09 & 19.99 & +29.21\\
      \hline
      {MUC\_MAS}~\cite{liumore}& 47.15 & 34.77 & 28.68 & 23.43 & +25.96\\
      \hline
      LwM~\cite{dhar2019learning} & 56.95 & 46.73 & 42.90 & 34.96 & +14.08 \\
      \hline
      LwF~\cite{li2017learning} & 70.83 & 61.20 & 51.83 & 44.96 & +2.26 \\
      \hline
      {MUC\_LwF}~\cite{liumore} & 69.55 & 59.30 & 53.12 & 47.76 & +2.03 \\
      \hline
      Ours & \textbf{71.12} & \textbf{62.25} & \textbf{54.75} & \textbf{49.73}&- \\
      \hline
      \end{tabular}
      }
  \end{center}
  \vspace{-5mm}
\end{table}

\begin{table}[t]
\scriptsize
\renewcommand\arraystretch{1.15}
  \begin{center}
    \caption{\small{Method comparison on Tiny ImageNet under the 5-phase setting. Top-1 classification accuracy is shown. Our system yields consistent improvement (by at least 2.61\%) shown by the column on the right.}}
    \vspace{-3mm}
    \label{5p_tiny}
    \setlength{\tabcolsep}{2.3mm}{
      \begin{tabular}{l|llll|c}
      \hline
      \multirow{2}{*}{Method} & \multicolumn{4}{c|}{Tiny ImageNet} & \multirow{2}{*}{\shortstack{All tasks avg.\\improvements}} \\
      \cline{2-5}
      \multirow{2}{*}{} & Task 2 & Task 3 & Task 4 & Task 5 & \multirow{2}{*}{} \\
      \hline
      Oracle & 65.62 & 64.73 & 59.10 & 57.63 & - \\
      \hline
      MAS~\cite{aljundi2018memory} & 35.23 & 24.40 & 19.59 & 14.53 & +21.27 \\
      \hline
      MUC\_MAS~\cite{liumore} & 37.10 & 26.97 & 23.39 & 18.24 & +18.28 \\
      \hline
      LwM~\cite{dhar2019learning} & 45.39 & 36.59 & 31.09 & 27.21 & +9.63  \\
      \hline
      LwF~\cite{li2017learning} & 53.40 & 42.17 & 32.94 & 25.70 & +6.15 \\
      \hline
      MUC\_LwF~\cite{liumore} & 52.90 & 45.33 & 38.05 & 32.09 & +2.61 \\
      \hline
      Ours & \textbf{56.83} & \textbf{48.17} & \textbf{39.35} & \textbf{34.46} &-\\
      \hline
      \end{tabular}
    }
  \end{center}
  \vspace{-5mm}
\end{table}

\begin{table*}[htb!]
\scriptsize
\renewcommand\arraystretch{1.15}
  \begin{center}
    \caption{\small{Top-1 accuracy comparison on CIFAR-100 under the 10-phase setting. We achieve superior results on all the 10 tasks. We outperform MUC\_LwF by 2.50\% on average.}}
    \vspace{-3mm}
    \label{10p_cifar}
    \setlength{\tabcolsep}{3.8mm}{
      \begin{tabular}{l|lllllllll|c}
      \hline
      \multirow{2}{*}{Method} & \multicolumn{9}{c|}{CIFAR-100} &  \multirow{2}{*}{\shortstack{All tasks average\\improvements}}  \\
      \cline{2-10}
      \multirow{2}{*}{} & Task 2 & Task 3 & Task 4 & Task 5 & Task 6 & Task 7 & Task 8 & Task 9 & Task 10 & \multirow{2}{*}{} \\
      \hline
      Oracle & 84.20 & 82.57 & 79.98 & 81.74 & 81.65 & 83.24 & 82.79 & 79.32 & 80.27 & - \\
      \hline
      MAS~\cite{aljundi2018memory} & 46.40 & 38.97 & 28.27 & 21.38 & 18.80 & 16.69& 13.62 & 12.29 & 11.71 & +24.42 \\
      \hline
      MUC\_MAS~\cite{liumore} & 48.75 & 40.87 & 32.20 & 26.30 & 23.35 & 24.29 & 19.86 & 18.97 & 21.27 & +19.11 \\
      \hline
      LwM~\cite{dhar2019learning} & 65.30 & 54.37 & 46.78 & 41.15 & 36.88 & 33.44 & 30.57 & 28.33 & 26.42 & +7.19 \\
      \hline
      LwF~\cite{li2017learning} & 69.00 & 61.07 & 52.52 & 46.46 & 40.78 & 37.04 & 32.04 & 30.24 & 26.56 & +3.58 \\
      \hline
      MUC\_LwF~\cite{liumore} & 65.15 & 57.43 & 50.95 & 45.28 & 42.87 & 41.56 & 36.29 & 33.84 & 32.08 & +2.50 \\
      \hline
      Ours & \textbf{69.20} & \textbf{61.50} & \textbf{54.40} & \textbf{49.52} & \textbf{45.80} & \textbf{41.67} & \textbf{37.80} & \textbf{35.32} & \textbf{32.70}  &  -  \\
      \hline
      \end{tabular}
      }
  \end{center}
  \vspace{-3mm}
\end{table*}

\begin{table*}[htb!]
\scriptsize
\renewcommand\arraystretch{1.15}
  \begin{center}
    \caption{\small{Top-1 accuracy on Tiny ImageNet under the 10-phase setting. Comparing with LwF and MUC\_LwF, the average improvement of our method is 2.15\% and 3.20\%, respectively.}}
    \label{10p_tiny}
    \vspace{-3mm}
    \setlength{\tabcolsep}{3.8mm}{
      \begin{tabular}{l|lllllllll|c}
      \hline
      \multirow{2}{*}{Method} & \multicolumn{9}{c|}{Tiny ImageNet} &  \multirow{2}{*}{\shortstack{All task average\\improvements}}  \\
      \cline{2-10}
      \multirow{2}{*}{} & Task 2 & Task 3 &Task 4 & Task 5 &Task 6 &Task 7 &Task 8 & Task 9 & Task 10 & \multirow{2}{*}{} \\
      \hline
      Oracle & 65.40 & 61.80 & 62.50 & 63.94 & 61.20 & 57.14 & 56.94 & 56.41 & 56.54 & - \\
      \hline
      MAS~\cite{aljundi2018memory} & 39.45 & 28.13 & 20.25 & 16.38 & 13.47 & 12.37 & 10.36 & 8.82 & 8.10 & +12.93 \\
      \hline
      MUC\_MAS~\cite{liumore} & 39.45 & 29.37 & 21.57 & 16.78 & 15.22 & 13.46 & 12.16 & 10.46 & 9.21 & +12.78 \\
      \hline
      LwM~\cite{dhar2019learning} & 46.55 & 40.10 & 35.97 & 30.36 & 27.90 & 23.28 & 21.11 & 19.35 & 18.84 & +1.14 \\
      \hline
      LwF~\cite{li2017learning} & 47.80 & 41.77 & 35.12 & 29.42 & 26.62 & 21.93 & 19.77 & 18.13& 13.84  & +2.15 \\
      \hline
      MUC\_LwF~\cite{liumore} & 42.35 & 37.77 & 31.57 & 29.04 & 26.83 & 22.29 & 21.00 & 18.64 & 15.47 & +3.20 \\
      \hline
      Ours & \textbf{48.25} & \textbf{41.97} & \textbf{36.50} & \textbf{31.30} & \textbf{28.78} & \textbf{24.27} & \textbf{22.57} & \textbf{20.91} & \textbf{19.17} &  -  \\
      \hline
      \end{tabular}
      }
  \end{center}
  \vspace{-3mm}
\end{table*}
\vspace{-2mm}
\subsection{Comparative Results}
In this section, we adopt three different incremental settings.
The three settings are 5, 10, and 6 phases.
5 and 10 phases mean that we split the data set into 5 or 10 tasks respectively. Each task has the same number of disjoint classes.
In the 5 phases setting, CIFAR-100 are divided into 5 equal parts. Each part has 20 classes. While for Tiny ImageNet, each part has 40 classes. 
In 10 phases setting, for CIFAR-100 and Tiny ImageNet, there are 10 and 20 classes in each task respectively.
6 phases is usually adopted by memory-based methods, which the first task is trained on half of the classes while the remaining classes are equally divided into 5 parts.
For the first task of CIFAR-100 and Tiny ImageNet, 50 and 100 classes are employed.
We run our whole learning process 3 times with different random seeds and report the average top-1 accuracy.

We compare our system with five state-of-the-art methods on memory-free class-incremental learning, \ie, MAS \cite{aljundi2018memory}, LwF \cite{li2017learning}, LwM \cite{dhar2019learning}, MUC\_MAS \cite{liumore}, MUC\_LwF \cite{liumore}, under 3 different incremental settings. Note that MUC\_MAS and MUC\_lwF require additional data to train auxiliary classifiers while ours does not. We fix the initial model for all methods and all setting.

\textbf{5-phase setting.}
For 5 phases, the top-1 accuracy of the initial model is 84.2 and 65.4 on CIFAR-100 and Tiny ImageNet respectively. Our results of the top-1 accuracy of subsequent tasks are shown in Table~\ref{5p_cifar} and Table~\ref{5p_tiny}. Our method outperforms all competitors with a clear margin.

\textbf{10-phase setting.}
Table~\ref{10p_cifar} and Table~\ref{10p_tiny} show the 10 phases setting results on CIFAR-100 and Tiny ImageNet. This setting is harder than 5 phases since the task sequence is longer and the first class number is smaller than 5 phases.
The top-1 accuracy of the first task is $83.7\%$ and $65.4\%$ for CIFAR-100 and Tiny ImageNet respectively. 
Our method outperforms MAS and MUC\_MAS by a large margin on two data sets. 
And for LwM and LwF on CIFAR-100, we gain $7.19\%$ and $3.58\%$ on all tasks average improvements respectively.
Note that MUC\_LwF that uses extra data sets, while we still beat it at all the tasks on all data sets.

\begin{figure}[!t]
\begin{center}
 \includegraphics[width=0.45\textwidth]{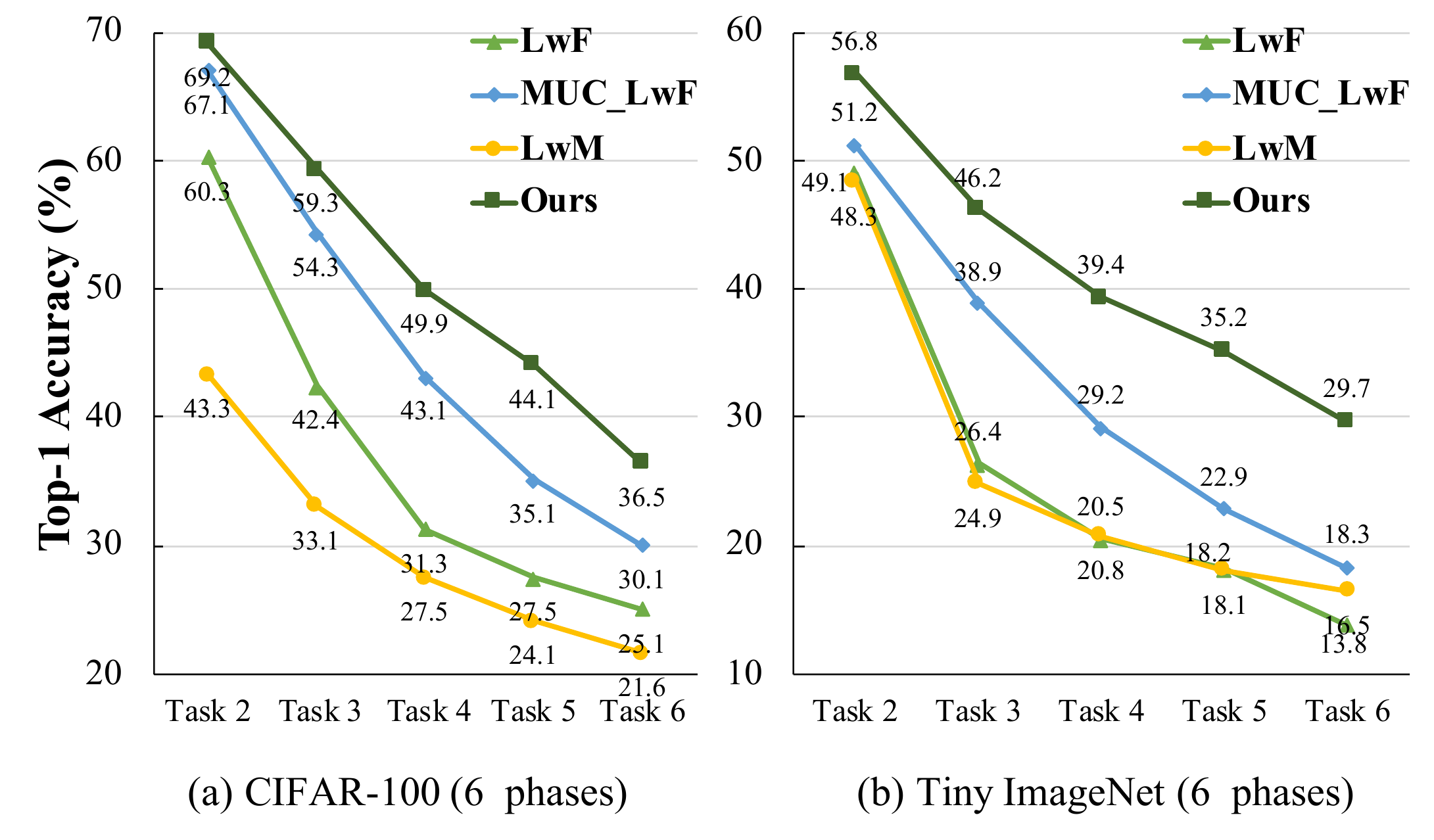}
 \end{center}
 \vspace{-5mm}
\caption{\small{Top-1 accuracy on CIFAR-100 and Tiny ImagNet under the 6-phase setting. We clearly observe the superiority of our method over the competing methods.}}
\label{6phases}
\vspace{-3mm}
\end{figure}

\textbf{6-phase setting.}
In addition, we employ 6 phases setting to evaluate the proposed method.
This setting is usually employed by memory-based method. We argue that this is because these methods require large number of classes to learn a model with strong generalization ability. Initialization with model like this is of great help to subsequent tasks. 
Fig.~\ref{6phases} shows the curve results that we compare with three methods on CIFAR-100 and Tiny ImageNet. Our method achieves the best performance among all competitors.

\vspace{-2mm}
\section{Conclusion}
In this paper, we analyze and tackle an inherent limitation of regularization-based class-incremental learning methods, \ie, catastrophic forgetting on fine-grained old classes. We make insightful analysis why it happens: the projections of new data on fine-grained old classes are not discriminative enough.
To solve this problem, we propose a memory-free generative replay strategy that provides direct image depictions of the fine-grained classes. By optimizing the generated images towards increasing diversity and reducing domain gap with old task images, we report noticeable improvement on the fine-grained classes and state-of-the-art accuracy in memory-free incremental learning.

{\small
\bibliographystyle{ieee_fullname}
\bibliography{egbib}
}

\end{document}